\documentclass[journal,a4paper]{IEEEtran}
\IEEEoverridecommandlockouts
\usepackage{cite}
\usepackage{amsmath,amssymb,amsfonts}
\usepackage{algorithmic}
\usepackage{graphicx}
\usepackage{textcomp}
\usepackage{xcolor}
\usepackage[a4paper, total={184mm,239mm}]{geometry}
\def\BibTeX{{\rm B\kern-.05em{\sc i\kern-.025em b}\kern-.08em
    T\kern-.1667em\lower.7ex\hbox{E}\kern-.125emX}}
\renewcommand{\baselinestretch}{0.98}

\usepackage[detect-none,binary-units=true]{siunitx}
\DeclareSIUnit{\nothing}{\relax}
\usepackage{multirow}
\usepackage{booktabs}
\usepackage{gensymb}
\usepackage{xcolor}
\usepackage{color} % color text in table
\usepackage{hyperref}
\hypersetup{%
    pdfborder = {0 0 0}
}
\begin{document}

\title{
Bio-inspired Autonomous Exploration Policies with CNN-based Object Detection on Nano-drones
}

\author{
\IEEEauthorblockN{
Lorenzo Lamberti\IEEEauthorrefmark{1}, 
Luca Bompani\IEEEauthorrefmark{1},
Victor Javier Kartsch\IEEEauthorrefmark{1},
Manuele Rusci\IEEEauthorrefmark{2},
Daniele Palossi\IEEEauthorrefmark{3}\IEEEauthorrefmark{4}
Luca Benini\IEEEauthorrefmark{1}\IEEEauthorrefmark{4}
}

\IEEEauthorblockA{\IEEEauthorrefmark{1} Department of Electrical, Electronic and Information Engineering, University of Bologna, Italy}

\IEEEauthorblockA{\IEEEauthorrefmark{2} Department of Electrical Engineering, KU Leuven, Belgium}

\IEEEauthorblockA{\IEEEauthorrefmark{3} Dalle Molle Institute for Artificial Intelligence, USI-SUPSI, Switzerland}

\IEEEauthorblockA{\IEEEauthorrefmark{4} Integrated Systems Laboratory, ETH Z\"urich, Switzerland}

Contact authors: lorenzo.lamberti@unibo.it,
luca.bompani5@unibo.it
% victorjavier.kartsch@unibo.it,\\
% manuele.rusci@kuleuven.be,
% daniele.palossi@idsia.ch,  %dpalossi@iis.ee.ethz.ch
% luca.benini@unibo.it,
}

\pagenumbering{gobble}
\markboth{This paper has been accepted for publication at IEEE Design, Automation, and Test in Europe (DATE) 2023.}{}
\maketitle

\IEEEaftertitletext{\vspace{-1\baselineskip}}

\begin{abstract}
Nano-sized drones, with palm-sized form factor, are gaining relevance in the Internet-of-Things ecosystem. 
Achieving a high degree of autonomy for complex multi-objective missions (e.g., safe flight, exploration, object detection) is extremely challenging for the onboard chip-set due to tight size, payload ($<$\SI{10}{\gram}), and power envelope constraints, which strictly limit both memory and computation.
Our work addresses this complex problem by combining bio-inspired navigation policies, which rely on time-of-flight distance sensor data, with a vision-based convolutional neural network (CNN) for object detection.
Our field-proven nano-drone is equipped with two microcontroller units (MCUs), a single-core ARM Cortex-M4 (STM32) for safe navigation and exploration policies, and a parallel ultra-low power octa-core RISC-V (GAP8) for onboard CNN inference, with a power envelope of just \SI{134}{\milli\watt}, including image sensors and external memories. 
The object detection task achieves a mean average precision of 50\% (at \SI{1.6}{frame/\second}) on an in-field collected dataset.
We compare four bio-inspired exploration policies and identify a pseudo-random policy to achieve the highest coverage area of 83\% in a $\sim$\SI{36}{\meter\squared} unknown room in a 3 minutes flight.
By combining the detection CNN and the exploration policy, we show an average detection rate of 90\% on six target objects in a never-seen-before environment.
\end{abstract}

\section*{Supplementary material}
In-field experiment video at: \url{https://youtu.be/BTin8g0nyko}.
Code at: \url{https://github.com/pulp-platform/pulp-detector}.

%  CONTENT
\section{Introduction} \label{sec:introduction}

The rapid evolution of the Internet of Things (IoT) is fueling the advent of flexible edge nodes powered by artificial intelligence (AI).
In this context, AI-based nano-sized unmanned aerial vehicles (UAVs), with a form factor of only $\sim$\SI{10}{\centi\meter} in diameter, can become revolutionary \textit{ubiquitous} smart sensing IoT nodes capable of exploring an environment while interacting with it in full autonomy~\cite{palossi2017self}, i.e., fulfilling the mission without the need of external resources.
The tiny and lightweight design of these pocket-sized drones provides enhanced safety for close-proximity human-machine interactions~\cite{frontnet}, as well as for indoor scenarios in narrow spaces, such as rescue missions~\cite{UAV_safety_rescue}. 

\begin{figure}[t]
\centering
\includegraphics[width=1\linewidth]{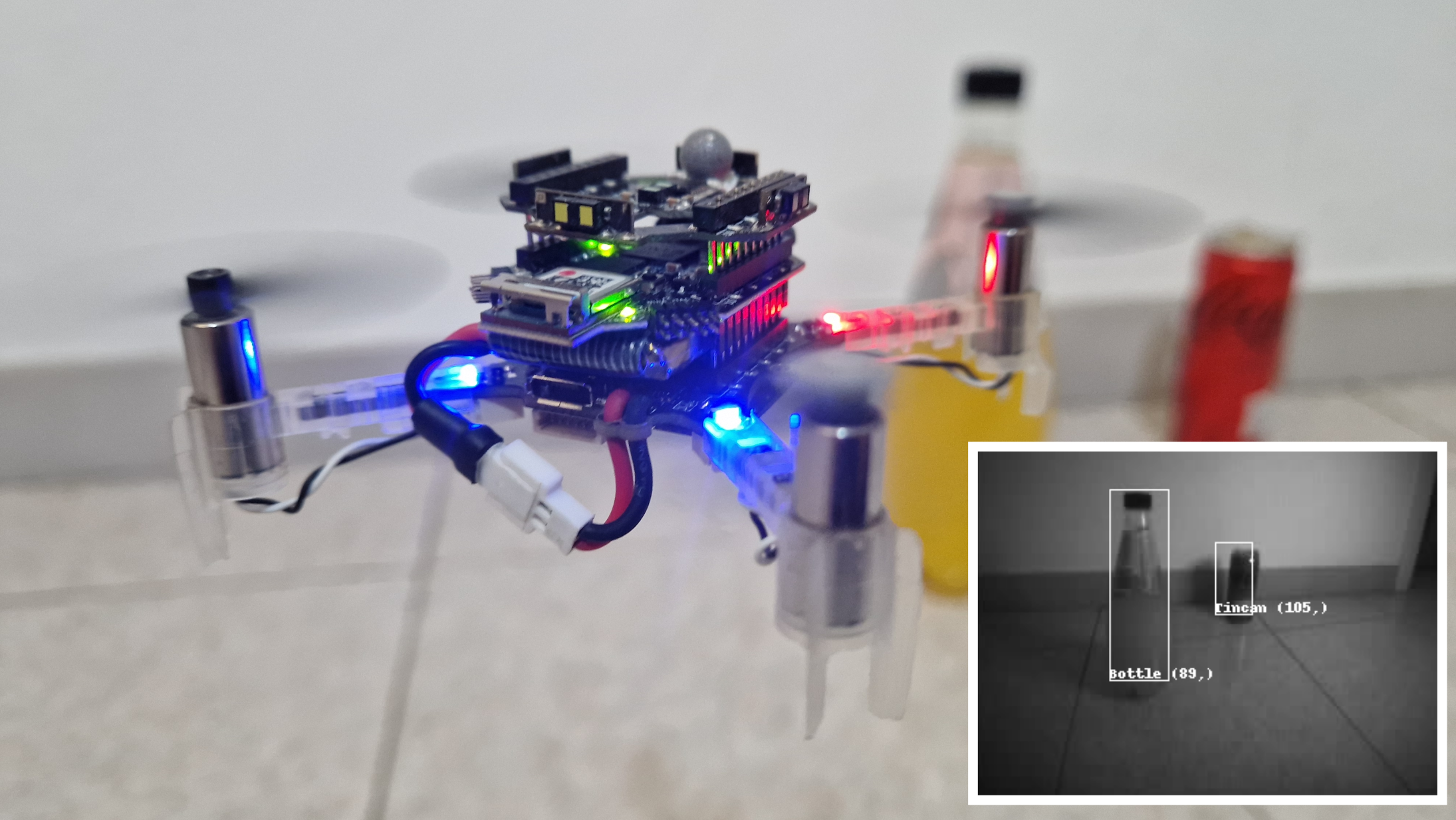}
\caption{Our fully autonomous prototype, based on a Crazyflie nano-drone.}
\label{fig:intro}
\end{figure}

Tackling such complex use cases requires the drone to reach a high level of intelligence typical of biological systems, which are autonomous and capable of handling multiple concurrent tasks~\cite{honeybees_visual_navigation} that span from basic control functionality to high-level perception~\cite{bio_multitask_GIURFA, webb2019internal}.
While autonomous nano-UAVs can already handle multiple basic functionality blocks at run-time (e.g., state estimation, low-level control), they still struggle to pursue multi-objective missions (e.g., safe navigation, exploration, visual recognition), which require computationally intensive multi-tasking perception~\cite{UAV_safety_rescue}.
In fact, with their ultra-constrained form factor and payload (a few tens of grams), these robots are prevented from hosting high-density/capacity batteries and bulky sensors or processors.
This scenario is even further exacerbated by the fraction of the total power budget allotted for the onboard electronics, i.e., 5-10\% of the total, which restricts the onboard processors to low-power microcontroller units (MCUs)~\cite{Wood2017}.

Inspired by the functional mapping of the different regions of human brains~\cite{dinstein2007brain}, we address this problem by mapping multiple tasks on the two MCUs available aboard our nano-drone.
The targeted system design follows the \textit{host-accelerator} template.
We leverage \textit{i.)} a single-core STM32F405 MCU, with a peak performance $<$\SI{100}{\mega\nothing} multiply-accumulate (MAC) operations per second,
for the execution of lightweight workloads (control, sensor-interfacing tasks), and \textit{ii.)} a parallel ultra-low-power (PULP) multi-core co-processor, GAP8, for coping with convolutional neural network (CNN) workloads ($\sim$\SI{1}{\giga MAC/\second}).

In this work, we contribute with a fully autonomous system design that combines bio-inspired exploration policies with a vision-based CNN for object detection aboard a nano-drone -- our prototype is shown in Fig.~\ref{fig:intro}.
We thoroughly characterize four bio-inspired exploration policies which use the ranging measures of multiple single-beam Time-of-Flight (ToF) sensors and run on the STM32F405 MCU.
Additionally, we deploy the object detection CNN on a GAP8 System-on-Chip (SoC) aboard our nano-drone, and we analyze its precision, computational/power costs, and performance by varying its depth.

Our in-field experiments show a peak performance for the biggest (deepest) object detector up to \SI{1.6}{frame/\second}, which recognizes two classes of objects with a mean average precision (mAP) score of 50\% within only \SI{134}{\milli\watt}.
On average, the best exploration strategy covers 83\% of a $\sim$\SI{36}{\meter\squared} room in a time budget of \SI{3}{\minute}.
Under the same testing environment, our closed-loop system recognizes six objects (two classes) with a mean detection rate of 90\% (5 independent runs) with a mean flight speed of \SI{0.5}{\meter/\second}.
Ultimately, to the best of our knowledge, we present the first fully autonomous nano-drone tackling a multi-objective mission, which consists of exploring, preventing collision, and detecting objects in real-time while relying only on onboard sensory and computational resources.

\section{Related works} \label{sec:related-works}

\textbf{AI-powered nano-drones.}
Recently, several AI-based visual pipelines have been brought to nano-drone systems. 
\textit{Lamberti et al.}~\cite{tiny_dronet} developed a CNN for end-to-end autonomous visual navigation that achieves an inference throughput of \SI{160}{FPS} on a nano-drone, thanks to a reduced CNN complexity (down to \SI{1.5}{\mega MAC}). 
This network design has two outputs: the steering angle and the collision probability. 
The latter is trained to detect any obstacle without providing information about its class or position within the image frame.
Similarly, \textit{NanoFlowNet}~\cite{nanoflownet} is a CNN for dense optical flow that aims at the obstacle avoidance task. It outputs per-pixel information about the obstacles in the scene, yet it does not give any information about their class.
\textit{Palossi et al.}~\cite{frontnet} deployed a CNN for human pose estimation aboard a nano-drone, allowing a drone to follow the movement of a human subject. 
However, this network is not designed to recognize multiple subjects in the same frame or different classes of subjects.
In contrast, our work aims to deploy an object detector to a nano-drone to detect multiple instances and classes of objects in its field of view. 

\textbf{Autonomous exploration}. 
Computer vision-based navigation techniques, based on complex feature extraction pipelines~\cite{CVPR13} or simultaneous localization and mapping (SLAM) techniques~\cite{SLAMBench2}, are reliable for autonomous robotic navigation. 
Still, as SLAM-based approaches have large memory requirements and rely on computationally intensive algorithms, they are exclusive to large UAV systems carrying heavy and high-power embedded computing systems~\cite{SLAMBench2}.

The state-of-the-art autonomous exploration approaches for nano-drones, which suffer from limited computational power and memory, take advantage of bio-inspired (or bug-inspired) algorithms, which rely on lightweight state-machine-based algorithms and low-power sensor readings\cite{kimberly_bug_inspired}.
For example, \cite{xu2014randombug} compares two bio-inspired exploration policies: the first changes the heading direction randomly after detecting an obstacle, while the second follows the obstacle's boundaries. 
Similar to the second one, \cite{kimberly_bug_inspired} implements a policy that follows the walls of a room, called wall-following, while \cite{spiral} describes a spiral motion.
Notable examples are described in~\cite{xu2014randombug}: an exploration policy either changes the heading direction randomly after detecting an obstacle or it follows the obstacle's boundaries, namely wall-following~\cite{kimberly_bug_inspired} and spiral~\cite{spiral}.
Moreover, there is a subcategory of bug-inspired algorithms that uses ranging measurements for navigating towards a target point~\cite{xu2013vectorization, kamon1996new}, even in the presence of obstacles. 
In our work, we adapt such bio-inpired ranging-based algorithms to our flying nano robotic platform and, for the first time, we integrate them with a visual object detection pipeline, evaluating the effectiveness of the exploration policy within a search mission. 

\textbf{Object detection on constrained MCUs.}
Among the CNN-based visual object detection pipelines, single-shot detector (SSD)~\cite{SSD} is a popular approach, consisting of a feature extraction backbone and multiple convolutional detection heads. 
To reduce the computation and memory costs, shallow backbones have been proposed, e.g., Mobilenet, while preserving the detection scores on widely used dataset~\cite{mobilenetv2}.

Focusing on the porting of object detectors on embedded systems, \textit{Tran Quang Khoi et al.}~\cite{SSD_RASPBERRY} deploy an SSD network (SSDLite-MobileNetV2) with a similar architecture as the one used in this paper onto a RaspberryPi B3+ mounted on a standard-sized drone.
The model execution reaches \SI{0.71}{FPS}, lower than our obtained throughput. 
More importantly, this solution exceeds both the power (up to \SI{3}{\watt}) and size constraints of the nano-drone system considered in this work.
\textit{Lamberti et al.}~\cite{lamberti_object_detection_gap} presented an SSD algorithm for license plates detection on a static multi-core MCU node (GAP8), achieving up to \SI{1.6}{FPS} with a power consumption of \SI{117}{\milli\watt}. 
Starting from this previous work, we design and integrate an SSD-based object detection CNN onto a nano-drone capable of exploring the environment while detecting specific objects. 
\section{System design} \label{sec:system}

\begin{figure*}[tb]
\centering
\includegraphics[width=1.0\linewidth]{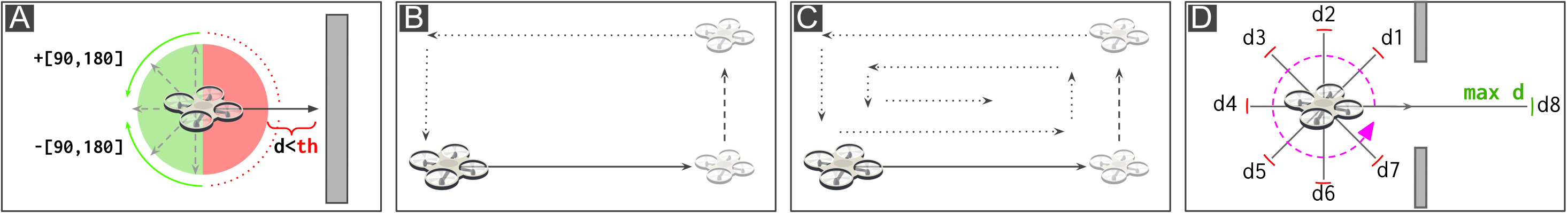}
\caption{Bio-inspired Exploration policies: 
(A) \textit{Pseudo-random}: inverts the direction by a random angle.
(B) \textit{Wall-following}: keeps a fixed distance from the perimeter.
(C) \textit{Spiral}: progressively increases the distance from the perimeter.
(D) \textit{Rotate-and-measure}: moves along the longest free-space direction.}
\label{fig:exploration_algorithms}
\end{figure*}

\subsection{Robotic platform} \label{sec:robotic_platform}

Our robotic platform revolves around Bitcraze Crazyflie 2.1 quadrotor, a commercial off-the-shelf (COTS) nano-drone with a weight of \SI{27}{\gram} and a diameter of \SI{10}{\centi\meter}. 
This quadrotor features an STM32F405 MCU that runs state estimation and actuation controls.
We extend this platform with three additional Bitcraze's COTS printed circuit boards (PCBs): the Flow deck, the Multi-ranger deck, and the AI-deck.
The Flow deck provides optical flow and height measurements used to increase the state estimation reliability.
The Multi-ranger deck features 5 single-beam VL53L1x ToF distance sensors mounted on the drone's top and sides, providing line of sight distance measurements within [0,4] meters range at \SI{20}{\hertz}.

Last, the AI-deck is a visual engine for nano-drones that extends the onboard processing capabilities of the Crazyflie.
This PCB embeds the GAP8 SoC, a general-purpose multi-core parallel ultra-low-power (PULP) processor.
GAP8 features a Custer of 8 RISC-V cores that can run up to \SI{175}{\mega\hertz}; it is designed for managing computation-intensive workloads with high energy efficiency, thanks to the parallel programming paradigm.
The on-chip memory is organized hierarchically, having \SI{64}{\kilo\byte} of L1 scratchpad memory shared among the 8 cores of the CL, and \SI{512}{\kilo\byte} of L2 SRAM memory off-the-cluster.
Furthermore, the AI-deck includes a low-power QVGA resolution grayscale camera (Himax HM01B0) and additional off-chip memories, including a \SI{8}{\mega\byte} of HyperRAM and a \SI{64}{\mega\byte} HyperFlash.

\subsection{Multi-task integration.} \label{sec:integration}

We exploit our platform to explore an environment while searching for specific objects.
To this aim, we separate the problem into two tasks: an \textit{exploration task}, targeting navigation inside an unknown mission area while avoiding collisions, and an \textit{object detection task} that runs the visual object detection SSD algorithm, as described in Sec.~\ref{sec:objdet}.
The exploration task runs a bio-inspired policy to determine the next set-point to feed the flight controller. 
As detailed in Sec.~\ref{sec:exploration_policies}, we study four different policies, i.e., the \textit{exploration algorithms}, which take the ranging measurements of the ToF sensors as input to prevent collisions.
On the other side, the performance for the \textit{object detection task} is accounted by measuring the successful detections of target objects present in the mission area.

The \textit{exploration} and the \textit{object detection} tasks do not require interaction with each other; therefore, we map them on the two MCUs aboard our nano-drone for concurrent execution.
More in detail, we use the two MCUs in a \textit{host-accelerator} configuration.
The STM32 is the \textit{host} and takes care of the flight controller and the exploration policy based on a lightweight state machine. 
The exploration policies use the ToF measurements from the Multi-ranger deck to determine the next set-point of the drone in terms of forward speed and yaw rate.
On the other hand, GAP8 is the \textit{accelerator} in charge of executing the most computationally demanding CNN algorithm.
This task includes camera acquisition and CNN processing and outputs the image-frame coordinates of the objects detected.
As a result, the processing is done fully onboard, making the final closed-loop system completely autonomous, i.e., with no external communication or computation.

\begin{table}[b]
\centering
\caption{Mean Average Precision (mAP) of the SSD CNNs trained on Google OpenImages and finetuned on the Himax dataset.}
\label{tab:ssd_results}
\begin{tabular}{cccccc} 
\toprule
\multirow{2}{*}{\begin{tabular}[c]{@{}c@{}}\textbf{Testing}\\\textbf{dataset}\end{tabular}} & 
\multirow{2}{*}{\textbf{Fine-tuning}} & 
\multirow{2}{*}{\textbf{Format}} & \multicolumn{3}{c}{\textbf{SSD size}} \\ 
\cmidrule{4-6}
 & & & 1$\times$ & 0.75$\times$ & 0.5$\times$ \\ 
\midrule
OpenImages & \textit{no} & \texttt{float32} & 59\% & 47\% & 43\% \\
Himax & \textit{no} & \texttt{float32} & 50\% & 41\% & 29\% \\
Himax & \textit{yes} & \texttt{float32} & 55\% & 46\% & 43\% \\
Himax & \textit{yes} & \texttt{int8} & 50\% & 48\% & 32\% \\
\bottomrule
\end{tabular}
\end{table}

\begin{figure*}[tb]
\includegraphics[width=1\linewidth]{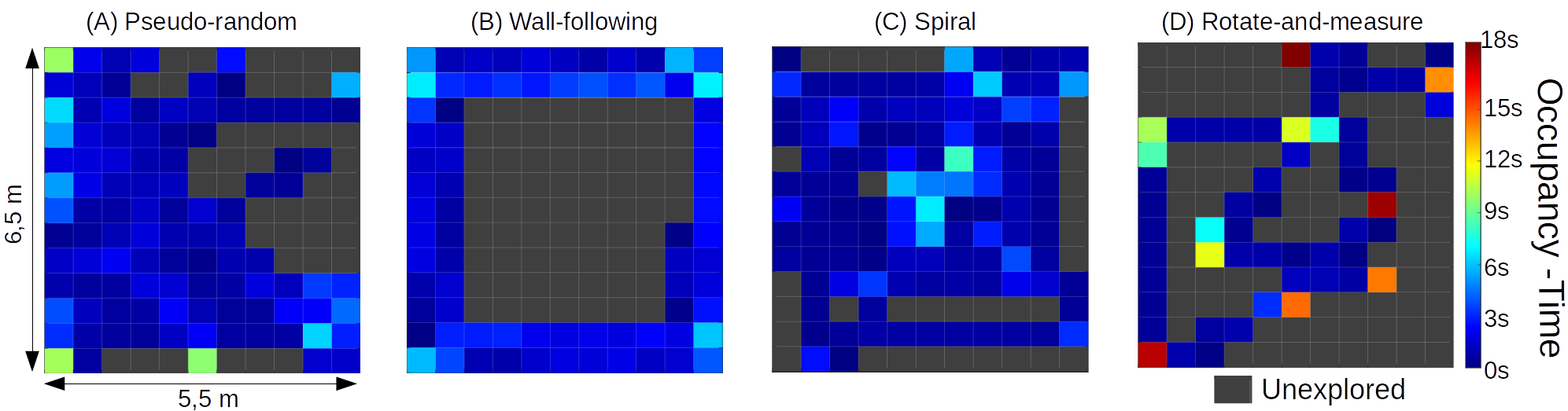}
\caption{Occupancy maps measured using the four exploration policies. During the experiments, the nano-drone flies at \SI{0.5}{\meter/\second}. The color of any cell, which represents a $0.5\times$\SI{0.5}{\meter} area, indicates the time spent by the nano-agent in that area (up to 18 sec). Black is used if the cell has not been explored.}
\label{fig:occupancy_map}
\end{figure*}

\subsection{Exploration algorithms} \label{sec:exploration_policies}

Our four bio-inspired exploration algorithms are shown in Fig.~\ref{fig:exploration_algorithms}.
All the policies rely on the ranging measurements from three ToF sensors: the front, left, and right ones. 
We assess the performance of these policies by tracking the drone's movements in a \SI{6.5}{\meter}$\times$\SI{5.5}{\meter} room equipped with a motion-capture system, tracking at \SI{50}{\hertz}. 
By discretizing the room's area into cells of $0.5\times$\SI{0.5}{\meter}, we plot the heatmap (Fig.~\ref{fig:occupancy_map}) representing the occupancy time of the drone over a \SI{3}{\minute} flight.
Our four exploration policies are:

\textbf{A) Pseudo-random.} This approach imitates the pseudo-random movement of biological creatures~\cite{xu2014randombug, webb2019internal}. 
The drone flies in a straight line as long as the ToF sensor does not identify obstacles within \SI{1}{\meter}. 
When an obstacle is identified, the drone rotates to a random value, which is always greater than $\pm$\SI{90}{\degree} from the current heading (Fig.~\ref{fig:exploration_algorithms}-A) to reduce the likelihood of detecting an obstacle previously identified.

\textbf{B) Wall-following.} This algorithm explores the perimeter of the room following its walls at a constant distance of \SI{0.5}{\meter}, measured by the side ToF sensors (Fig.~\ref{fig:exploration_algorithms}-B). 
The navigation stops when a front obstacle is detected and resumes after a $\pm$\SI{90}{\degree} turn towards an obstacle-free heading.
By construction, this algorithm never explores the inner part of the testing room.

\textbf{C) Spiral.} The drone first explores the environment by performing several concentric perimetral paths (such as wall-following), each with an ever-increasing distance from the external walls (Fig.~\ref{fig:exploration_algorithms}-C).
Once the room's center has been reached, the process is reversed, i.e., the consecutive perimetral explorations are performed with an ever-decreasing distance to the walls. 
The process starts over once the drone has reached its initial position.
This spiral exploration starts with an initial distance from the walls of \SI{0.5}{\meter}, and, for each lap, it is increased/decreased by the same amount.

\textbf{D) Rotate-and-measure.} This algorithm features two sequential phases (Fig. ~\ref{fig:exploration_algorithms}-D). 
In the first phase, the drone scans the environment by performing a \SI{360}{\degree} spin in place while measuring the front distance every \SI{45}{\degree}.
In the second phase, the drone flies towards the most obstacle-free direction for a maximum of \SI{2}{\meter}.
This policy favors exploring the inner areas of our testing room while frequently neglecting its corners.

\subsection{Object detection algorithm design} \label{sec:objdet}

Our object detection pipeline is based on an SSD algorithm composed of a MobilenetV2 feature extractor~\cite{mobilenetv2, object_detection}, pre-trained on the full OpenImages~\cite{openimages} dataset, and multiple detection heads, which is the ending part of the CNN that predicts the locations, categories, and confidence scores of objects at different resolution scales. 
To trade off latency/memory and detection accuracy, we deploy three different CNNs by varying the width multiplier $\alpha$ of the MobilenetV2 backbone.
We refer to the different object detection algorithms as \textit{SSD-MbV2-$\alpha$}, where $\alpha=\{0.5,0.75,1.0\}$. 
The largest \textit{SSD-MbV2-1.0} model features \SI{4.67}{\mega\nothing} parameters and requires \SI{534}{\mega MAC} operations while the \textit{SSD-MbV2-0.75} and \textit{SSD-MbV2-0.5} have, respectively, \SI{2.68}{\mega\nothing} and \SI{1.34}{\mega\nothing} parameters and require \SI{358}{\mega MAC} and \SI{193}{\mega MAC} operations.

\begin{figure}[b]
\centering
\includegraphics[width=1\linewidth]{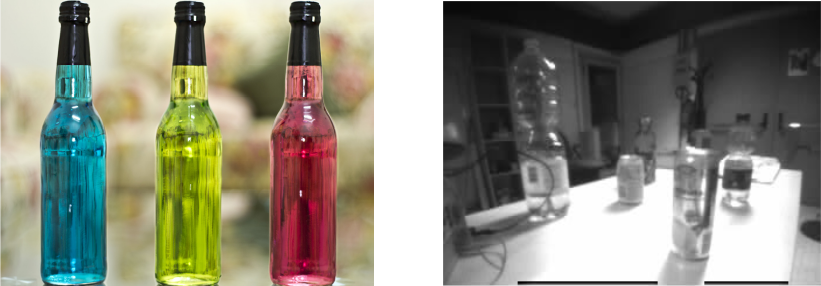}
\caption{Google OpenImages sample (left) vs. onboard sensor image (right).}
\label{fig:comp}
\end{figure}

We train our SSD models to detect two object categories, \textit{bottles} and \textit{tin cans}, using a subset of images from the OpenImages collection.
Given the unbalanced training set, i.e., 1306 and 11306 images for tin cans and bottles, respectively, we balance the dataset by generating additional tin can images through horizontal translation (up to 10\% of the image's width).
The final split consists of 19142 images for training, 208 for validation, and 663 for testing, where the validation and testing portions match the original dataset splitting.
Finally, to overcome the domain shift between the training and real-world data (Fig.~\ref{fig:comp}), we add a fine-tuning phase on an additional dataset we collected and called \textit{Himax Dataset}, which includes 321 training images and 279 testing images.

To deploy the object detector on GAP8, we quantize the CNN to 8-bit.
We perform an additional fine-tuning step of quantization aware training (QAT) to minimize the mAP loss due to the 8-bit conversion.
Then, we exploit the QAT routine included within the Tensorflow Object Detection API framework~\cite{object_detection} using symmetric quantization ranges because the software primitives of the target hardware rely on symmetric integer ranges.
Lastly, Greenwaves's GAP\textit{flow} toolset\footnote{https://greenwaves-technologies.com/gapflow/} is used to produce the C code of the object detection task, constraining the L2 buffer size to \SI{250}{\kilo\byte}.

\section{Results} \label{sec:results}

\subsection{Object detection evaluation} \label{sec:obj_detector_eval}

The SSD models are trained on OpenImages for 1200 epochs using the RMSProp optimizer. 
We set a learning rate of $8 \cdot 10^{-4}$ with an exponential decay of $0.95$ every 24 epochs and a batch size of 24.
The OpenImages frames are resized to $320\times240$ pixels to match the resolution of the drone onboard camera. 
During training, the images are extended with photometric augmentations, such as flipping, brightness adjustment, random cropping, and grayscale conversion, individually applied with a probability of 0.5.
Before deployment, the SSD models are fine-tuned, also applying QAT, on the Himax dataset for 100 epochs, with a learning rate of $10^{-4}$ retaining the same exponential decay of $0.95$ every 10 epochs.

Tab.~\ref{tab:ssd_results} reports the mAP scores (as defined for the COCO dataset~\cite{COCO_api}) of the multi-sized SSD models on both the OpenImages and our Himax testing datasets.
The largest model achieves an mAP of 59\% when trained and tested on images from the OpenImages dataset, up to 16\% vs. the smallest model. 
These scores reduce by 9\%, 6\%, and 14\% for the three models when tested on the Himax dataset. 
We argue this effect to be due to the onboard camera image quality being lower than the web-retrieved images.
After fine-tuning on the Himax training set, the mAP score improves up to 55\% for the best model, bridging the accuracy gap for the mAP score on the OpenImages test set. 
Given the best scores of 50\% and 48\% after 8-bit quantization, we consider both \textit{SSD-MbV2-1} and \textit{SSD-MbV2-0.75} models for the in-field evaluation.

Tab.~\ref{tab:ssd_on_board} reports the performance of the SSD CNN running on the GAP8 SoC. 
We set the operating voltage at \SI{1.2}{\volt} and a clock frequency of \SI{160}{\mega\hertz} for the multicore cluster, while the peripheral clock is set to \SI{250}{\mega\hertz}. 
The largest models (SSD-MbV2-1.0) can process up to \SI{1.6}{frame/\second}, with a computational efficiency of 5.3 \textit{MAC/clock cycles}.
Instead, the smaller SSD-MbV2-0.75 and SSD-MbV2-0.5 result, respectively, 1.6$\times$ and 2.7$\times$ faster than the larger model.
The power consumption of the AI-deck reaches a peak of \SI{143.5}{\milli\watt} when running the SSD-MbV2-0.75 model, where the inference task shows the highest compute efficiency, maximizing the memories' bandwidth and processing logic utilization. 
Conversely, the power consumption decreases to \SI{134.5}{\milli\watt} if running the most accurate SSD-MbV2-1.0 model.

\begin{table}
\centering
\caption{SSD CNNs' onboard performance.}
\label{tab:ssd_on_board}
\begin{tabular}{ccccc} 
\toprule
\textbf{SSD} & \textbf{Parameters} & \textbf{Operations} & \textbf{Efficiency} & \textbf{Throughput}  \\ 
\midrule
1$\times$    & 4.7M                & 534 MMAC            & 5.3 MAC/cycles      & 1.6 FPS              \\
0.75$\times$ & 2.7M                & 358 MMAC            & 5.9 MAC/cycles      & 2.3 FPS              \\
0.5$\times$  & 1.2M                & 193 MMAC            & 5.3 MAC/cycles      & 4.3 FPS              \\
\bottomrule
\end{tabular}
\end{table}

\begin{figure}[b]
\includegraphics[width=1\linewidth]{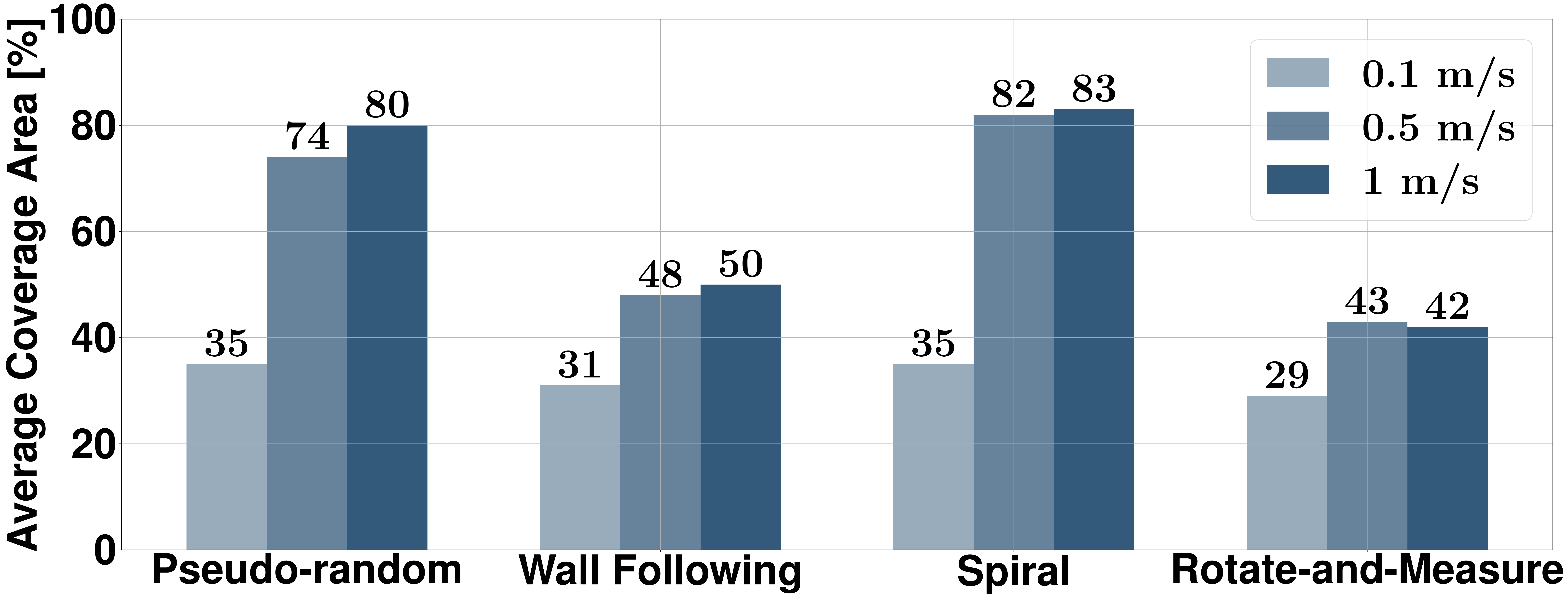}
 \caption{Average coverage area (in \%) for each exploration policy, varying its mean flight speed (i.e., \SI{0.1}{\meter/\second}, \SI{0.5}{\meter/\second}, and \SI{1}{\meter/\second}).}
\label{fig:occupancy_bars}
\end{figure}

\subsection{Exploration policies evaluation} \label{sec:results_exploration}

To assess the performance of our four exploration policies, we calculate the \textit{coverage area} (\%) in the testing room described in~\ref{sec:exploration_policies} by calculating the ratio between the visited cells w.r.t. their total (i.e., 143 cells).
We mark a cell ``visited'' when the drone's center of mass falls into it.
We evaluate each exploration policy with three average flight speeds (i.e., \SI{0.1}{\meter/\second}, \SI{0.5}{\meter/\second}, and \SI{1}{\meter/\second}), obtaining 12 test configurations.
For every configuration, we perform five runs of \SI{3}{\minute} each, accounting for a total of 60 runs (\SI{3}{\hour} flight time).

Fig.~\ref{fig:occupancy_bars} reports the coverage area of each configuration (policy and speed) averaged on the five runs.
The \textit{pseudo-random} and \textit{spiral} policies are those that mainly benefit from higher speeds, passing from $35\%$ coverage area (at \SI{0.1}{\meter/\second}) to $74\%$ and $82\%$ (at \SI{0.5}{\meter/\second}), and finally to $80\%$ and $83\%$ (at \SI{1}{\meter/\second}), respectively.
Instead, the \textit{wall-following} and \textit{rotate-and-measure} exploration policies show a slight improvement passing from a mean flight speed of \SI{0.1}{\meter/\second} to \SI{0.5}{\meter/\second}, as much as $+17\%$ and $+14\%$, respectively.
Contrary to the previous policies, the highest flight speed does not improve the coverage area, i.e., $+2\%$ and $-1\%$, respectively.
The \textit{wall-following} policy has a limited performance since the drone only explores the room's perimeter (see Fig.~\ref{fig:occupancy_map}-B).
% For the \textit{wall-following}, the main reason for its limited performance comes from the drone only exploring the perimeter of the room (see the occupancy map in Fig.~\ref{fig:occupancy_map}-B).
Similarly, the \textit{rotate-and-measure} policy spends the vast majority of the \SI{3}{\minute} flight spinning in place and focusing on the center of the room (see Fig.~\ref{fig:occupancy_map}-D).

\subsection{In-field closed-loop system evaluation} \label{sec:results_infield}

To evaluate the closed-loop in-field performance, we measure the detection rate of each policy/CNN by placing three bottles and three tin cans in the room, relying on the same testing configuration of Sec.~\ref{sec:results_exploration}: $6.5\times$\SI{5.5}{\meter} testing room, \SI{3}{\minute} flights.
One bottle and one tin can are close to the center, while the other four are near the corners. 
The final detection rate depends on \textit{i.)} detector's precision and throughput and \textit{ii.)} covered area (partially depending on the flight speed). 
In Tab.~\ref{tab:detection_rate}, we evaluate the best two SSD CNNs (see Sec.~\ref{sec:obj_detector_eval}) with all four exploration policies and three flight speeds.
The bigger \textit{SSD-MbV2-1.0} consistently achieves, for any configuration, an equal or greater detection rate than the medium 0.75$\times$ model, suggesting that, in our setup, the detector's accuracy is more important than its throughput (\textit{SSD-MbV2-0.75} has higher throughput than \textit{SSD-MbV2-1.0}, but lower mAP).

\begin{table}[t]
\centering
\caption{Average detection rate: 6 objects, 5 runs of 3 minutes each.}
\label{tab:detection_rate}
\begin{tabular}{cccccc} 
\toprule
\multicolumn{1}{l}{\multirow{2}{*}{\textbf{SSD }}} & \multirow{2}{*}{\begin{tabular}[c]{@{}c@{}}\textbf{Flight}\\\textbf{speed [m/s]}\end{tabular}} & \multicolumn{4}{c}{\textbf{\textbf{Detection rate}}} \\ 
\cmidrule{3-6}
\multicolumn{1}{l}{} & & \begin{tabular}[c]{@{}c@{}}Pseudo \\random\end{tabular} & \begin{tabular}[c]{@{}c@{}}Wall \\following\end{tabular} & Spiral & \begin{tabular}[c]{@{}c@{}}Rotate and\\measure\end{tabular} \\ 
\midrule
\multirow{3}{*}{1.0$\times$} & 0.1 & 27\% & \textbf{63\%} & 67\% & \textbf{53\%} \\
 & 0.5 & \textbf{90\%} & 50\% & \textbf{73\%} & \textbf{53\%} \\
 & 1 & 83\% & 53\% & 70\% & 47\% \\ 
\midrule
\multirow{3}{*}{0.75$\times$} & 0.1 & 27\% & \textbf{50\%} & 33\% & 47\% \\
 & 0.5 & \textbf{80\%} & 37\% & \textbf{43\%} & \textbf{50\%} \\
 & 1 & 37\% & 27\% & \textbf{43\%} & 33\% \\
\bottomrule
\end{tabular}
\end{table}

\begin{figure}[b]
\includegraphics[width=1\linewidth]{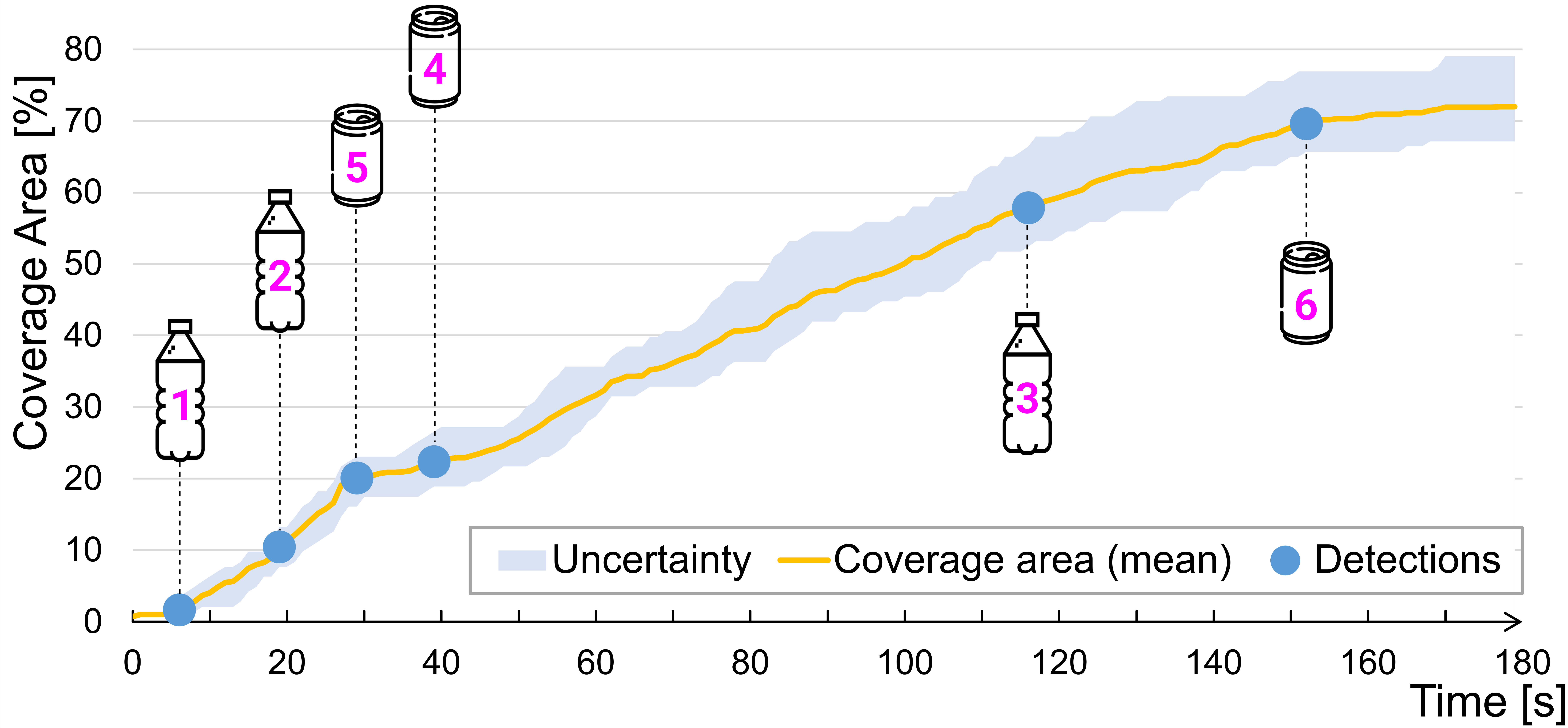}
\caption{Coverage area of the \textit{random} policy over 5 runs (mean and variance). The detection time of the 6 target objects (blue dots) refers to one single run.}
\label{fig:final_system}
\end{figure}

Focusing on the \textit{SSD-MbV2-1.0}, the peak performances are achieved by the \textit{pseudo-random} and \textit{spiral} policies, which show the highest detection rate at the intermediate flight speed of \SI{0.5}{\meter/\second}. 
In contrast, in Sec.~\ref{sec:results_exploration}, the best coverage area was obtained by the highest flight speed, which suggests \SI{1}{\meter/\second} being too high for the limited inference rate of the bigger 1$\times$ SSD (i.e., \SI{1.6}{frame/\second}).
Then, the \textit{wall-following} and \textit{rotate-and-measure} policies show the lower detection rates, $63\%$ and $53\%$ at most, respectively.
This behavior is the consequence of the limited coverage area of the policies, where the former misses all the objects placed in the center of the room while the latter marginally explores the perimeter. 

In Fig.~\ref{fig:final_system}, we focus our analysis on the best performing model: \textit{pseudo-random} exploration policy with the nano-drone flying at \SI{0.5}{\meter/\second} and running the \textit{SSD-MbV2-1.0}.
The yellow line shows the coverage area over five \SI{3}{\minute} flights, which scores a maximum average of 72\% coverage, with a variance of 21\%.
The blue dots indicate the time each object is recognized for a run achieving 100\% detection rate in \SI{154}{\second}.

\begin{table}[tb]
\centering
\caption{Power breakdown of the robotic platform.}
\label{tab:power_table}
\begin{tabular}{cccccc} 
\toprule
\multicolumn{1}{c}{} & Motors & CF elect. & AI-deck & Multi-ranger & \textbf{Total}\\ 
\midrule
Power [W] & 7.32 & 0.277 & 0.134 & 0.286 & \textbf{8.02}\\
Percentage & 91.31\% & 3.45\% & 1.67\% & 3.57\% & \textbf{100\%}\\
\bottomrule
\end{tabular}
\end{table}

Lastly, Tab.~\ref{tab:power_table} shows the power breakdown of the nano-drone platform, measured by profiling each component individually with the Power Profiler Kit 2 by Nordic Semiconductor.
The power cost of the AI-deck, running the biggest \textit{SSD-MbV2-1.0}, accounts for 1.67\% of the total power, which is dominated by the motors (91.31\%).
The other electronics components, i.e., Crazyflie's MCU and the ToF sensors, require the remaining 7.03\% of the total power. 
\section{Conclusions} \label{sec:conclusions}

This work presents the first multi-objective autonomous nano-drone, pursuing the exploration of an unknown environment while avoiding collision and searching objects with a vision-based CNN detector.
Our work compares four bio-inspired exploration policies and three versions of object detectors running independently on two resource-limited MCUs aboard the nano-drone. 
The best configuration reaches a final detection rate of 90\%, exploiting \textit{i.)} a pseudo-random policy for exploration, \textit{ii.)} the largest object detection model, and \textit{iii.)} a mean flight speed of \SI{0.5}{\meter/\second}.
This result shows how the higher detection rate can be reached by trading off the detection capabilities of the CNN, its throughput, and the mean flight speed of our nano-drone.

\section*{Acknowledgments}
We thank the Center for Research on Complex Automated Systems and Aurora Di Giampietro for their support.
This work has been partially funded by the Autonomous Robotics Research Center of the UAE Technology Innovation Institute.

\bstctlcite{IEEEexample:BSTcontrol}

\bibliographystyle{IEEEtran}
\bibliography{IEEEabrv,bibliography}

% that's all folks
\end{document}